# Kaggle's Kinship Recognition Challenge: Introduction of Convolution-Free Model to boost conventional ensemble classifier

TENG Guangway          TIAN Mingchuan          BAO Yipeng

## 1. Introduction

Convolutional neural networks (CNNs) have achieved remarkable success in computer vision, making them a dominant classifier for ensemble learning in numerous Kaggle competitions. Nevertheless, this work aims to explore a convolution-free base classifier that can be used to widen the variations of the conventional ensemble classifier. Specifically, we propose Vision Transformers as base classifiers to combine with CNNs for a unique ensemble solution in Kaggle's kinship recognition. The idea behind the proposed solution is based on the belief that: If we achieve a lower correlation between base classifiers by using a particular method, such as by introducing a different model which calculates the prediction using different methods, the final ensemble classifier can achieve a greater boost. In this paper, we verify our proposed idea by implementing and optimizing variants of the Vision Transformer model on top of the existing CNN models. The combined models achieve better scores than conventional ensemble classifiers based solely on CNN variants. We demonstrate that highly optimized CNN ensembles publicly available on the Kaggle Discussion board can easily achieve a significant boost in ROC score by simply introducing variants of the Vision Transformer model to the ensemble because of its low correlation magnitude between the CNNs in this experiment. In addition, we also conclude that the convolution-free Vision Transformer may even be used as an additional booster classifier to an already highly optimal ensemble of only CNN models to further boost the competitor's ranking.

## 2. Background

### 2.1 Problem Statement of the Kaggle competition

The Kaggle competition 'Recognizing Faces in the wild' is a kinship recognition challenge officially held by Northeastern SMILE lab in 2019. The goal of this challenge is to build a model by determining if two individuals are biomedically related based solely on images of their faces. Since the dataset is relatively imbalanced (most of the individuals are not related), each submission is evaluated on areas under the Receiver Operating Characteristic (ROC) curve between the predicted probability and the observed target. The ROC is commonly used to benchmark the prediction results for Kaggle's binary classification problems, simply because the calculation is based on the true positive and the false positive rate which are non-related to the distribution of the dataset.

The dataset is given as the images of different individuals, people in the same family were collected and put in the same directory, but not every individual is related to each other in the same family. For example: The husband and the wife do not share a kinship relationship with each other. To provide competitors with training instances of image pairs of two related individuals, a file named train_relationship.csv of 3,598 related pairs was provided. The training dataset contains 326 families and 2,085 different people. The competitors need to predict the kinship relationship of 5,310 pairs of individuals in the test dataset defined in  sample_submission.csv with the true label being withheld. One of the difficulties of the task is that predicting kinship relationships only through facial recognition has its limitation. People from different families may have similar facial features. individuals that share kinship relationships may also have various different facial features too. Therefore, even the best model may misjudge kinship relationships. Refer below for the images of the same person as shown in Figure 1. One can note that the images provided may contain other people not related to the person of interest, his/her images may also have different color bases, or different ages, which makes it harder to extract similar features out of the individual.

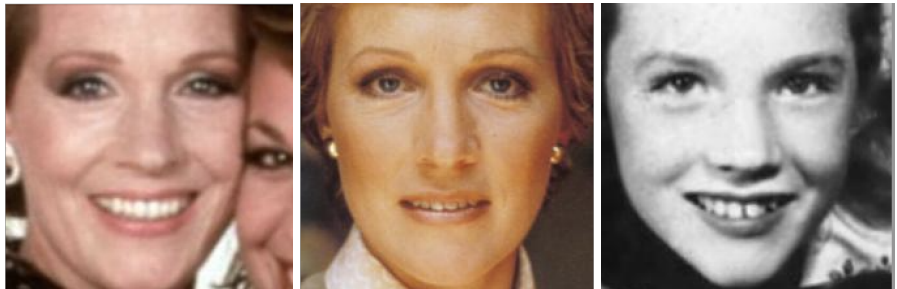

Figure 1 Images of the same person

## 2.2 Related methodology and limitation

Based on observations of the discussion board, top competitors revolve around the idea of implementing a vast variant of Siamese Neural network setup based on VGGFace-Resnet50 architecture and a few less common CNN models. The solutions of all the variants were combined to achieve a highly optimal ensemble classifier output for submission.

As a standalone model, the VGGFace-Resnet50 has the greatest ROC score of all models discussed on the public board. The best publicly released kernel, as a standalone model, can achieve ROC scores of 0.893 and 0.900 in the public and private leaderboards respectively. However, the majority of the variants created have relatively high correlations as they are fundamentally similar to each other (Convolution neural network based). Therefore this technique may require many variants to combine to derive at a very good solution

For instance, upon the end of the Kaggle competition, the champion with username 'mattemillio', shared in the public discussion board how he derived his final solution by combining 37 solutions to achieve first place by combining a wide range of CNN model solutions. In his opinion there are three methods to create new CNN variants.

1. Concatenating different combinations of features produced by VGGFace results in new models that are not correlated.
2. Combining features outputs of VGGFace and Facenet to generate a third model based on the other two CNNs used for combination.
3. Changing structure of fully-connected layers after concatenation of features

Eventually, he created 30 models exhaustively based on the three methods listed above to combine with 7 public models as his final ensemble classifier, achieving his best score of 0.923 and 0.917 for private and public ROC respectively.[13]

# 3. Proposed method

## 3.1 Motivation of the proposed method

The motivation for this proposed method came from our observation that the competitors active on the public discussion board revealed how reliant they were on the VGGFace-Resnet50 CNN algorithm to make a comprehensive ensemble classifier. This is due to the lack of other kinds of standalone models that can be implemented successfully and yield very high ROC scores during the competition 3 years ago.

Therefore, to widen the variations of the ensemble's base classifiers output, we propose to introduce the Convolution-Free Vision Transformer model into the application of this category of Kaggle competition. The model has not been discussed in the public board and it was very unlikely to be used by any of the top competitors during the competition 3 years ago. The Convolution-Free Vision Transformer model algorithm calculates the prediction result differently when compared to conventional CNN models used in this Kaggle competition. When combined together with CNN models into one ensemble classifier, this shall allow us to achieve a lower correlation magnitude between the base classifiers within the ensemble classifier.

We shall verify through experiments that we may combine a more moderate number of the lower correlated Transformer model predictions with deep CNN model predictions, to achieve a ROC score very close to the champion, "mattemillio" who used 37 deep CNN models to win the competition.

## 3.2 Pyramid Vision Transformer (PVT)

We chose the Pyramid Vision Transformer (PVT) as our Transformers model for ensemble learning. PVT was proposed by Wang [1] as an improved version of Vision Transformer by having two major advantages: 1) It generates multi-scale feature maps between each block like the CNNs for dense prediction and better interpretability, 2) It introduces the Spatial-reduction technique which largely reduces the computational overhead. Simply put, the PVT was chosen because it requires less computation and memory to train each model compared with the traditional Vision Transformer. An overview of the PVT architecture is depicted in Figure 2.

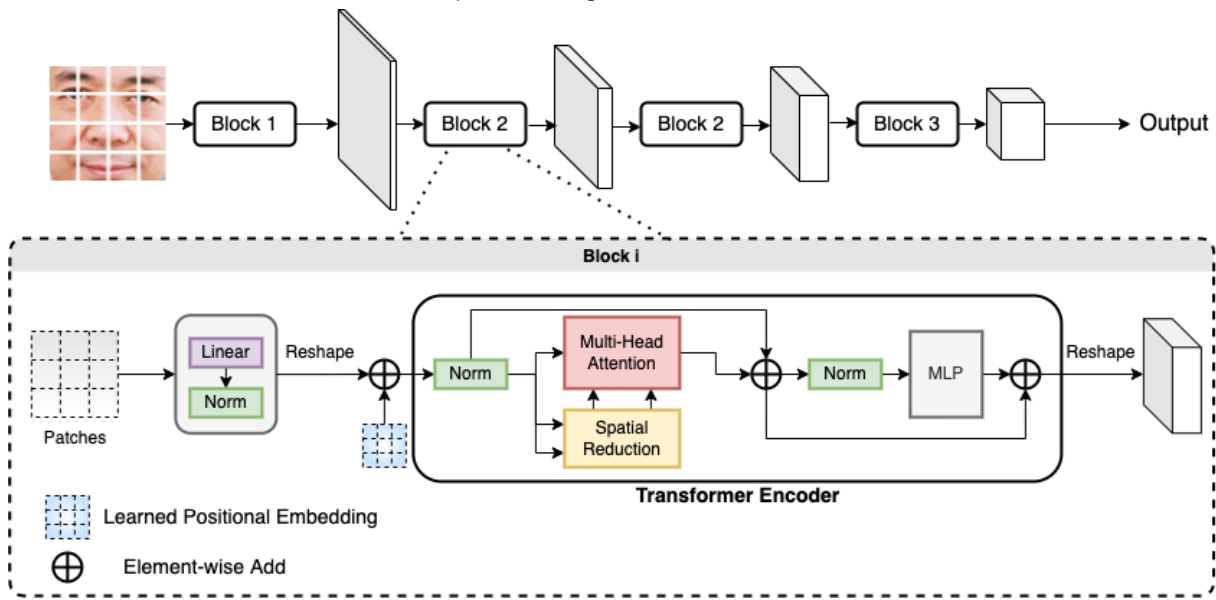

Figure 2: **Overall architecture of Pyramid Vision Transformer (PVT)**

In the PVT, the Transformer encoder in the stage $i$ has $L_i$ encoder layers, each of which is composed of an attention layer and a feed-forward layer. The PVT is easier to train because it replaces the traditional multi-head attention layer [2] with the spatial-reduction attention (SRA) layer. This newly designed SRA layer performs similarly to multi-head attention which receives a query Q, a key K, and a value V as input, and outputs refined features. However, the SRA reduces the spatial scale of K and V before the attention operation to reduce the computation overhead [1]. The detailed formulation of SRA is presented as follows:

$$Reduce(x) = Norm(Reshape(x, R_i)W^s)$$

The above formula describes how the spatial reduction is done to the input sequence. Here, $x$ represents an input sequence, $R_i$ denotes the reduction ratio, $W^s$ is the linear projection that reduces the dimension of input sequence. $Reshape(x, R_i)$ reshapes the input $x$ to the size $\frac{HW}{R^2} \times (R^2 C)$.

$$Attention(Q, K, V) = Softmax(\frac{qk^T}{\sqrt{d_{head}}})v$$

$$head_j = Attention(QW^Q, Reduce(K)W^k, Reduce(V)W^v)$$

$$SRA(Q, K, V) = Concat(head_0, ..., head_{Ni})$$

The rest of the computation is identical to the original multi-head attention. We first define the $Attention$ operation as the pairwise similarity between two elements of the sequence and their respective Q and K. Here, $W^Q, (K)W^k, W^v$ are linear projection parameters for Q, K, and V, respectively. $Ni$ is the head number of the attention layer at Stage $i$. Once the attention score is calculated for each head, they are concatenated for the final SRA output. Therefore, SRA is a simple but effective attention layer that is capable of processing high-resolution feature maps while reducing computational and memory costs. The final output of the PVT is a feature vector and can be fed into a Siamese Network for the downstream task of kinship verification.

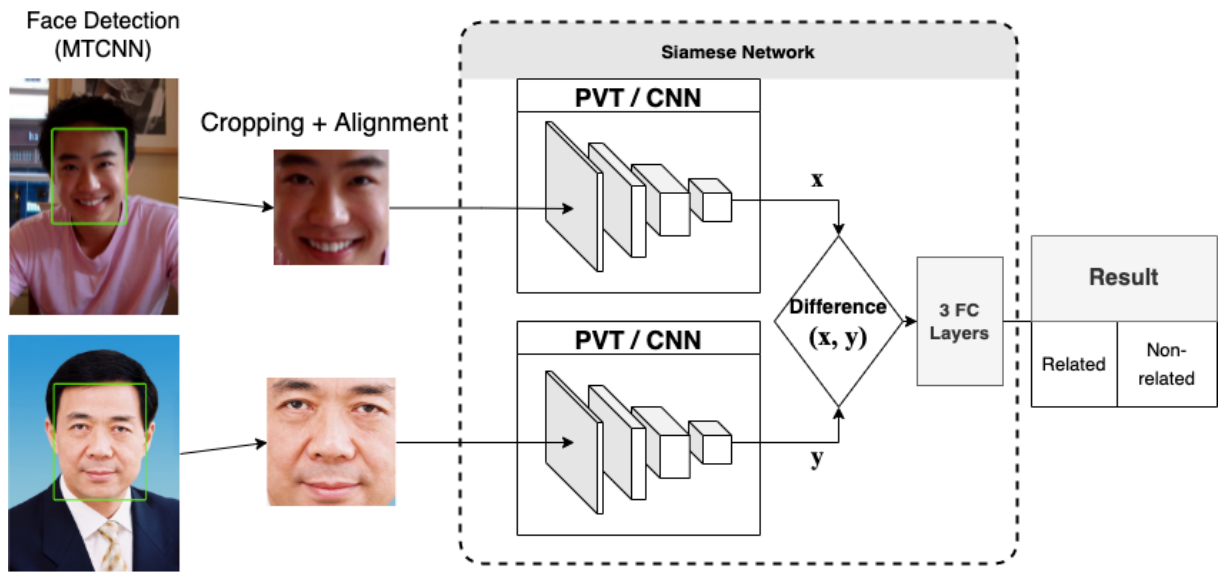

Figure 3: **Experiment pipeline – Siamese Neural Network**

# 4. Experiments

## 4.1 Vision Transformers

**Settings.** We choose PVT-Tiny [1] and PVT-v2-b0 [3] as our base model for pretraining. Because they are the smallest models (13.2M and 3.4M Params) among the variations of PVTs which still perform well according to their Top-1 accuracy on ImageNet-1K (70.5% and 75.1%) [4]. We select the CASIA-Webface [7] dataset for model training and use the Labeled Face in the Wild (LFW) [8] dataset for model testing. CASIA-Webface is an open dataset composed of 494,141 images classified into 10,575 identities. The LFW is a benchmark dataset to validate the performance of face recognition algorithms commonly seen in recent research papers. Note that both the CASIA-webface and LFW datasets are pre-processed using MTCNN [9] for face detection and alignment to ensure the same distribution of each image. Our pretrained PVT-Tiny and PVT-v2-b0 achieve 96.45% and 97.13% on the LFW dataset.

For kinship recognition, we adopt the Siamese Neural Network (SNN) [10] to construct an experiment pipeline as shown in Figure 3. Within the SNN, the two images are fed into two identical networks which result in two distinct feature outputs. The difference between the feature outputs is computed before it is fed into three fully-connected layers. Finally, we use Cross Entropy Loss to train our model weights and a SoftMax function before the final output. Using different quadratic combinations of the output features, we are able to construct four variants of Transformers as shown in Table 1.

Table 1: **PVT variants by modifying Difference(x, y)**

| Variant (Base model) | Difference (x, y) | Kaggle ROC Private / Public |
|---|---|---|
| PVT-1 (PVT-Tiny) | $x - y$ | 0.828 / 0.838 |
| PVT-2 (PVT-Tiny) | $Concat\,((x^2 + y^2), (x^2 - y^2), (x \cdot y))$ | **0.854 / 0.855** |
| PVT-3 (PVT-Tiny) | $Concat\,((x - y), (x - y)^2, (x^2 + y^2), (x^2 - y^2), (x \cdot y))$ | **0.847 / 0.860** |
| PVT-4 (PVT-v2-b0) | $Concat\,((x - y), (x - y)^2, (x^2 + y^2), (x^2 - y^2), (x \cdot y))$ | 0.815 / 0.821 |

We use the above combinations of quadratic functions because they converge faster and achieve better overall accuracy among all approaches we tried in the error-and-trial process. Note that for the kinship recognition dataset, faces are also detected and aligned using MTCNN to ensure the kinship image data have the same distribution as the CASIA-Webface data used for pretraining.

**Results.** In Table 1, we see that PVT variants achieve acceptable Public Scores from Kaggle, placing all of them to the top 20% of the leaderboard. Additionally, concatenating different combinations of features leads to models that are not fully correlated. In Figure 4, we see that PVT variants have diverse distributions of classification results on the kinship test dataset. Therefore, we see a great potential to use these PVT variants as additional classifiers to boost the performance of the ensemble model which has a long-existing problem of over-reliance on the CNNs.

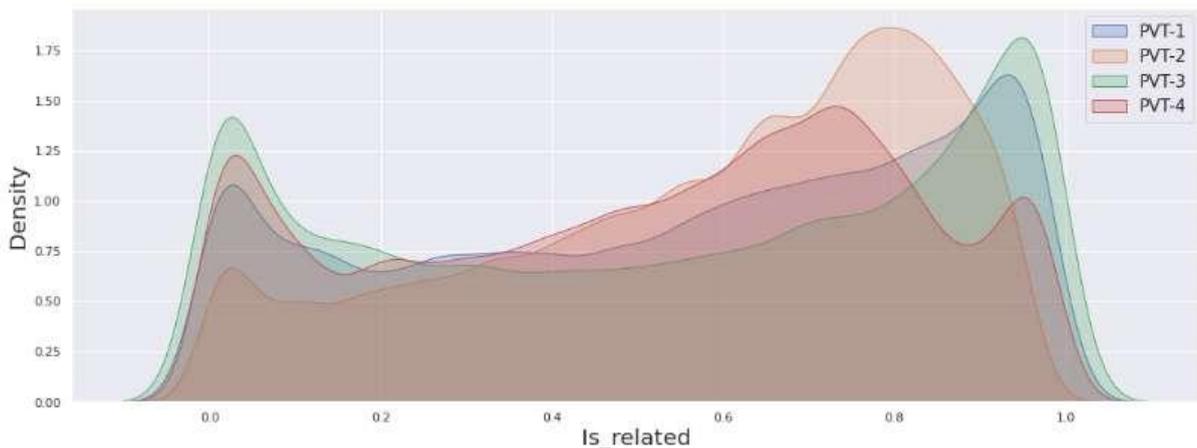

Figure 4: **Distribution of classification scores from different PVT variations**

## 4.2. CNNs

**Settings.** To retrieve prediction results from conventional CNN models, we use the source code provided by the public community of this Kaggle Competition. As the source code provided by the public notebooks was developed 3 years ago, some slight modifications were required to run the model. We train the dataset using the Kaggle notebook provided. Due to the time constraint of submission, we omitted fine-tuning all the models used in our experiment and settled with the prediction results once it is close to what the notebook author managed to achieve with it. We selected 5 notebooks to evaluate our proposed methodology. The majority of these models' architecture are VGGFace-Resnet50 CNN based variants, and other less common CNNs variants.

Results of all the CNN model prediction solutions produced in the experiment are shown in Table 2 and Table 3. The models mentioned in Table 2 were purposely minimally tuned with reduced epoch to obtain a weaker prediction result such that we can demonstrate the impact of the vision transformer combined into ensembles made of other weaker CNNs.

Table 2: **Minimally tuned CNN models**

| Model name | VGG1_unfully_trained | VGG2_unfully_trained |
|---|---|---|
| Kaggle ROC score (Private/Public) | 0.841/0.851 | 0.854/0.852 |

*Notes: VGG1_unfully_trained and VGG2_unfully_trained are trained models that were done hastily at reduced epochs to achieve a decent unoptimised performance. The model designs are similar to VGG-1 and VGG-2 mentioned in Table 3.*

Table 3: **Well-Optimized CNN Models based on hyperparameters relatively similar to default values set by notebook authors**

| Model name | VGG-1 | VGG-2 | VGG-3 | VGG-4 | VGG-5 |
|---|---|---|---|---|---|
| **Kaggle ROC score (Private/Public)** | 0.872/0.872 | 0.868/0.862 | 0.895/0.889 | 0.871/0.874 | 0.902/0.899 |

*Notes: The links to all the public notebooks are mentioned in the Appendix.*

**Results.** In Table 3 we see that the CNNs variants shared at the discussion board were also able to achieve ROC scores which surpass our Vision Transformer as a standalone prediction. We can note that in general the VGG variants also have relatively diverse distribution. The diversity of these VGG variants allows the competitors to achieve outstanding scores during the kaggle competition by combining all 5 of them together into a new ensemble. Figure 5 demonstrates the already wide diversity of the VGG and other CNN variants used in our experiment. As there is often a limit to how much uncorrelated VGG and other CNN variants can be created during the competition, we shall propose and show later in the setup that further combining using the PVT variants solution will achieve a greater ROC score.

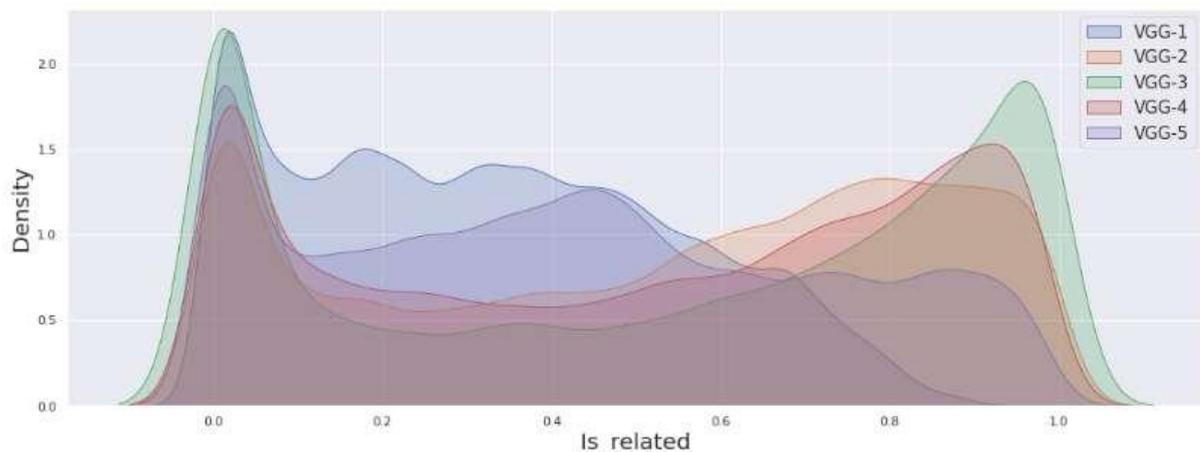

Figure 5: **Distribution of classification scores from different VGG variations**

## 4.3 Ensemble Classifier Model

Upon retrieving solutions from all the above models. We begin combining results based on the following combination to achieve the predictions of our Ensemble models. For the ensemble, we combined all solutions by the weight sum approach. Models that are uncorrelated and have higher ROC scores are given heavier weightage.

**Setting1.** We ensemble a moderately optimal transformer's solution, as per PVT-1 described in Table 1 and two decent VGGFace-Resnet50 solutions mentioned in Table 2 and named the prediction small_Final_ensemble. We then separately ensemble the two VGGFace-Resnet50 solutions mentioned in Table 2 without the PVT-1 transformer's solution and name the ensemble prediction as small_VGG_ensemble. We combine all solutions by the weight sum approach. Models that are uncorrelated and have higher ROC scores are given heavier weightage.

Finally, we compare and evaluate the difference between the two ensembles to assess how much improvement the Transformer has contributed to the ROC score of the small_VGG_ensemble in the case when it was added to the ensemble in the small_Final_ensemble.

Table 5: **Experiment 1 key setups result**

| Model name | small_VGG_ensemble | PVT-1 | small_Final_ensemble |
|---|---|---|---|
| **Kaggle ROC (Private/Public)** | 0.875/0.879 | 0.828/0.838 | 0.891(+0.016) /0.899 (+0.02) |

**Results.** In Table 5, we can see that even though the PVT-1 stand alone ROC score is significantly weaker than the small VGG ensemble, it managed to significantly boost the small decent VGG ensemble by a private, public score of 0.016 and 0.020 respectively. We believe that this was contributed by the low correlation magnitude between the introduced transformer solution and the other 2 VGG ensemble. This further verified our original thoughts that the PVT-1 output being lowly correlated would make an impactful booster when introduced into the CNN ensemble. We can further observe the diversity of the small_VGG_ensemble and PVT-1 in Figure 6. We compute the PCC correlations of all solutions in the setup as shown in Table 7.

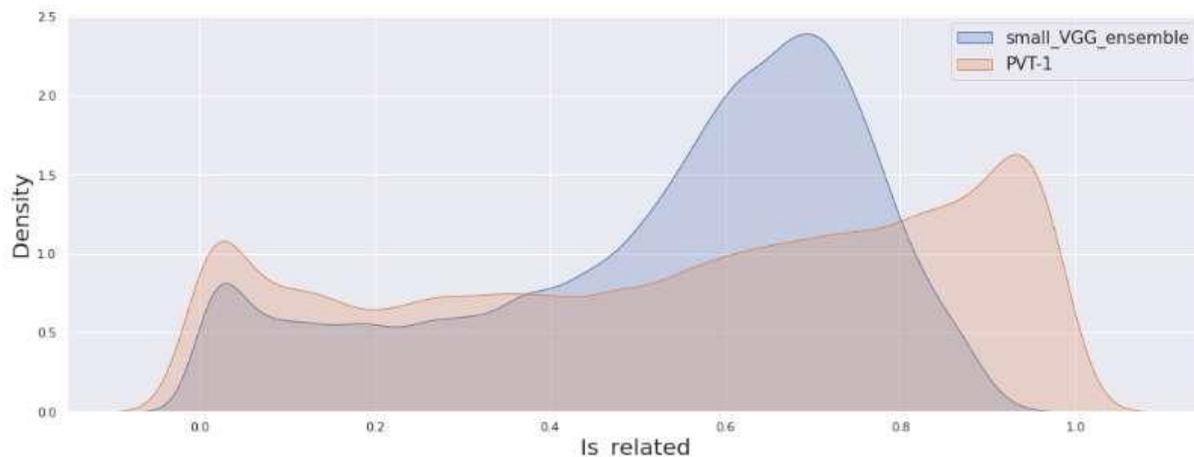

Figure 6: **Distribution of classification scores from small_VGG_ensemble and PVT-1**

**Setting2.** We ensemble all 4 of our best Transformer solutions mentioned in Table 1 and named it PVT_ensemble, then separately ensemble all 5 of our well optimized VGGFace-Resnet50 CNN solutions mentioned in Table 3 and named it VGG_ensemble. Finally, we ensemble the solution of PVT_ensemble and VGG_ensemble and named it Final_ensemble. The final prediction of the Final_ensemble was submitted to achieve our best ROC score of 0.914 / 0.913 (Private/Public leaderboard), ranking our score equivalent to the 11th and 8th place on the private and public leaderboard respectively. We also compare the ROC score of VGG_ensemble before it was combined with PVT_ensemble to assess how much contribution PVT_ensemble on top of the conventional VGG_ensemble has made to achieve our high score.

Table 6: **Experiment 2 key setups result**

| **Model name** | VGG_ensemble | PVT_ensemble | Final_ensemble |
|---|---|---|---|
| **Kaggle ROC (Private/Public)** | 0.912 / 0.908 | 0.862 / 0.874 | **0.914 (+0.002) / 0.913 (+0.005)** |

Table 7: **Experiment 2, Ensembles leaderboard ranking comparison**

|  | **VGG_ensemble Rank** | **Final_ensemble Rank** | **PVT_ensemble Ranking boosted** | **PVT_ensemble Score Boost** |
|---|---|---|---|---|
| Kaggle ROC (Private/Public) | 14th / 18th | 11th / 8th | 3 / 10 | + 0.002 / +0.005 |

**Results.** From Table 6, we can see that despite a more limited ROC score boost as compared to Setup 1, we were still able to boost our score sufficiently enough to increase our leaderboard ranking by +3/+16 private and public leaderboard respectively as shown in Table 7. The competitors' scores equivalent to our ensembles' are shown in Image 1 and Image 2 for private and public leaderboards respectively. The boost provides a +0.002/+0.005 private/public leaderboard score to the Final_ensemble. This is tremendous considering we have only implemented 4 PVT variants in PVT_ensemble.

Image 1: **Ensembles private leaderboard ranking comparison of Experiment 2**

| 11 | ▼ 1 | dsad | | Final_ensemble | 0.914 | 19 | 3Y |
| 12 | ▲ 4 | Andreisun | | | 0.913 | 134 | 3Y |
| 13 | ▲ 8 | Hong zio | | | 0.913 | 6 | 3Y |
| 14 | — | ethan.cheng | | VGG_ensemble | 0.912 | 126 | 3Y |

Image 2: **Ensembles public leaderboard ranking comparison of Experiment 2**

| 8 | Chaohui Song | | Final_ensemble | 0.913 | 32 | 3Y |
| 9 | dasdasd | | | 0.913 | 23 | 3Y |
| 10 | dsad | | | 0.913 | 19 | 3Y |
| 11 | billzxz | | | 0.913 | 51 | 3Y |
| 12 | 雨女无瓜 | | | 0.913 | 62 | 3Y |
| 13 | sajidjai | | | 0.913 | 30 | 3Y |
| 14 | ethan.cheng | | | 0.912 | 126 | 3Y |
| 15 | pi-null-mezon | | | 0.912 | 111 | 3Y |
| 16 | Andreisun | | | 0.911 | 134 | 3Y |
| 17 | BM_AI | | | 0.911 | 84 | 3Y |
| 18 | dzw_ | | VGG_ensemble | 0.908 | 195 | 3Y |

From Figure 7 below, we can also see that the 2 highly optimized ensembles still have a considerable amount of diversity to incorporate into each other's solution. This also further explains how we achieve that tiny boost even though both ensembles are already highly optimized.

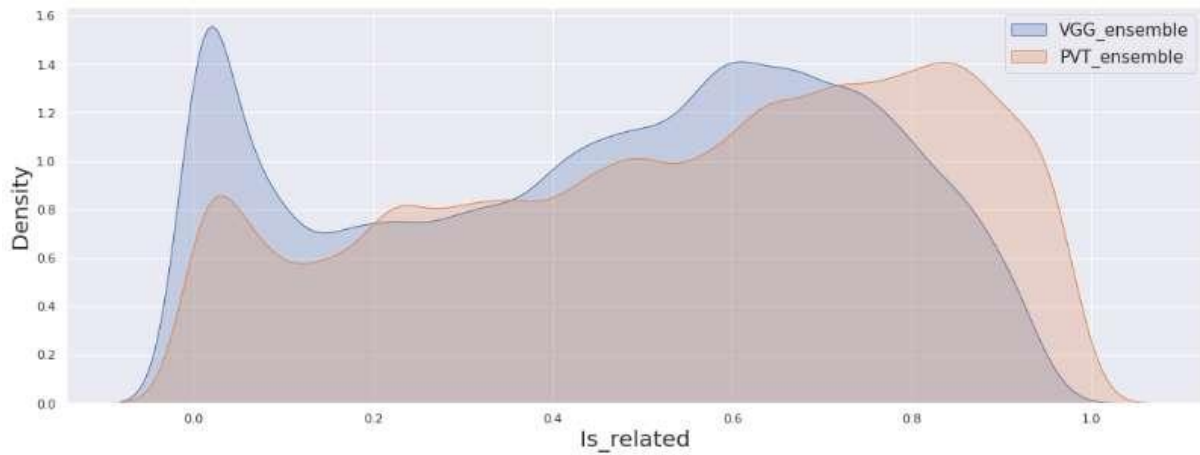

Figure 7: **Distribution of classification scores from final ensemble models**

The results of the above 2 setups are thoroughly elaborated and discussed in the following section of the paper.

## 5. Discussion

## 5.1 Pearson Correlation Coefficient (PCC) Analysis

The PCC Matrix of Experiment setup 1 solutions are as shown in Table 8. Here we can note that the PVT-1 model solution has the lowest correlation when compared to the conventional VGGFace-Resnet50 CNN variants made by public notebook authors to boost their ensemble classifiers. This denotes that the PVT model prediction is more effective to boost the ensemble classifier if given the same standalone performance score, explaining why we could receive so tremendous boost when combined with small_VGG_ensemble even though the ROC score of PVT-1 is much lower than small_VGG_ensemble and the standalone CNN models used as base classifier in this ensemble.

Table 8: **Correlation Matrix of all the base classifiers used in experiment 1 using Pearson Correlation Coefficient**

| Correlation Table of Experiment 1 | VGG-1_unfully_trained | VGG-2_unfully_trained | small_VGG_ensemble | PVT-1 |
|---|---|---|---|---|
| VGG-1_unfully_trained | 1 | 0.6496 | 0.8585 | **0.5971** |
| VGG-2_unfully_trained | 0.6496 | 1 | 0.9476 | **0.5467** |
| small_VGG_ensemble | 0.8585 | 0.9476 | 1 | **0.6197** |
| PVT-1 | **0.5971** | **0.5467** | **0.6197** | 1 |

The PCC Matrix of Experiment setup 2 solutions are as shown in Table 9. Here we can note again that the PVT_ensemble solution, which is based on the PVT models, has proven to have the lowest overall correlation when compared to the conventional VGGFace-Resnet50 CNN variants made by public notebook authors.

Table 9: **Correlation Matrix of all the base classifiers used in experiment 2 using Pearson Correlation Coefficient**

| Correlation Table of Experiment 1 | VGG-1 | VGG-2 | VGG-3 | VGG-4 | VGG-5 | VGG_ensemble | PVT_ensemble |
|---|---|---|---|---|---|---|---|
| VGG-1 | 1 | 0.67934 | 0.7161 | 0.7144 | 0.7817 | 0.8584 | **0.6771** |
| VGG-2 | 0.67934 | 1 | 0.7173 | 0.7086 | **0.7343** | 0.8667 | **0.6621** |
| VGG-3 | 0.7161 | 0.7173 | 1 | 0.7341 | 0.7914 | 0.9067 | **0.6941** |
| VGG-4 | 0.7144 | 0.7086 | 0.7341 | 1 | 0.7720 | 0.8864 | **0.7023** |
| VGG-5 | 0.7817 | **0.7343** | 0.7914 | 0.7720 | 1 | 0.9156 | 0.7489 |
| VGG_ensemble | 0.8584 | 0.8667 | 0.9067 | 0.8864 | 0.9156 | 1 | **0.7843** |
| PVT_ensemble | **0.6771** | 0.6621 | **0.6941** | **0.7023** | 0.7489 | **0.7843** | 1 |

## 5.2 Transformers Face Area Analysis

Interpretability of Vision Transformers has been a research focus in explainable AI (XAI). A common approach is to visualize the parts of the image that led to a certain classification either using attention maps or employing heuristic propagation along the attention graph. In our analysis, we experimented with a novel approach by Cherfer [5] published in CVPR2021, which computes the relevancy between the selected area with the image as a whole for Transformer models. Specifically, the method assigns local relevancy based on the Deep Taylor Decomposition [6] and then propagates these relevancy scores through the layers. We demonstrate that our PVT models attend to the face area as we expected, as shown in Figure 8. The areas highlighted in red have the highest relevance with the area in the box.

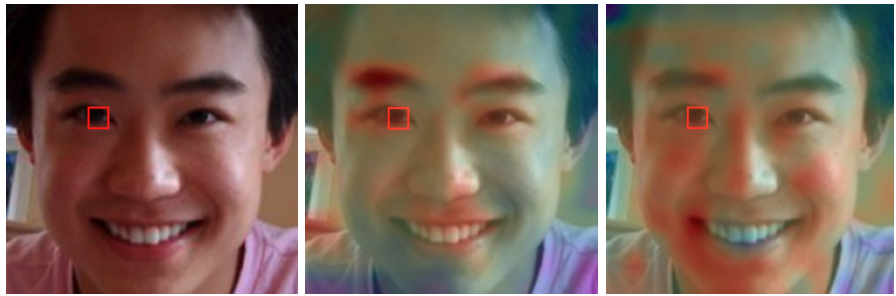

Figure 8. Area relevancy computed using Deep Taylor Decomposition for PVT-3

## 5.3 Final Results

The transformer model PVT has proven to be capable of being integrated into conventional ensemble classifiers used in this Kaggle competition to further raise ROC score and leaderboard position. In setup 1, weaker based CNN classifiers were able to receive a big ROC score boost of +0.012/+0.011, private/public scores by the combination of 1 additional PVT model solution that are also relatively weak compared to the other standalone model predictions at the public discussion board. In setup 2, the VGG_ensemble was made by a combination of 5 well optimized models. As such the ensemble ROC score before combining with the pvt_ensemble is already very high at 0.912/0.908. However the PVT model can still be combined to give a tiny boost of 0.002/0.005, raising the final_ensemble score to 0.914/0.913. The magnitude of the ROC score boost may be lesser than setup 1 case, but it is still worth implementing as doing so raises our score's final private leaderboard ranking from 14 to 11 and public leaderboard ranking from 18 to 8. (+3/+10)

To achieve the champion's high score of 0.923 / 0.917, one can simply create more variants of PVT and combine them with more CNNs on top of Setup 2. It is very likely that one may achieve the champion's hi-score with less than 37 variants as we can see from Table 9 that the PVT variants are generally less correlated and may be further optimized as an individual classifier. We abstained from challenging the champion directly as this may still be an extremely time consuming procedure.

# 6. Conclusions and Future work

We combine Vision Transformers with CNNs to formulate ensemble models for the Kaggle kinship recognition challenge. We demonstrate that Vision Transformers extract features differently from CNNs via the self-attention mechanism, resulting in less correlated prediction results. Extensive experiments verify that Vision Transformers and CNNs can produce distinct base classifiers which can be ensembled to obtain better predictive performance. Due to the time constraint, we are unable to ensemble more models as well as to fine-tune our solution by error-and-trial practice. However, our proposed method shows great potential by reaching the top-10 on the Kaggle leaderboard within the limited time frame.

Although the PVTs serve well as additional classifiers to the CNNs in ensemble models, there are still some specific modules and operations that are not considered in this work. In addition to the Cross-Entropy Loss used in this work, other loss criterions such as Contrastive Loss [11] and Triplet Loss [12] may add to our SNN to leverage the label information more effectively during the training. Vision Transformers is still in its early stage of development by any means. Therefore, we believe there are many potential variants to be explored in the future, and hope that our PVT-CNN ensemble model could serve as a good starting point.

# 8. Appendix

## 8.1 Team Contribution

TIAN Mingchuan
- Adopted the Pyramid Vision Transformer for our downstream task.
- Pretrained Transformer models on large public face dataset, fine-tuned on target dataset.
- Constructed a Siamese Neural Network and tested various combinations of feature outputs.
- Conducted attention area analysis based on Deep Taylor Decomposition [5, 6].

BAO Yipeng
- Pre-processed image data
- Ran CNN experiment
- Help with ensemble models
- Fine-tuned transformer

TENG Guangway
- Trained CNNs used in the experiment by repairing broken codes on the kaggle public discussion board and running them.
- Performed model ensemble of Vision Transformer outputs and CNNs
- Tabulated ROC scores of our experiment for both ensemble and standalone based classifier
- Evaluated the relationship between ensemble performance and the PCC matrix and its performance as an individual classifier.

## 8.2 Kaggle Leaderboard

Our final result in Kaggle: Private:0.914, Public 0.913

**Public Leaderboard (April 24th)**

**Private Leaderboard (April 24th)**

## 8.2 Public Kaggle Notebook Referenced

1. VGG-1 Public Notebook link:
   https://www.kaggle.com/code/castep/vggface
2. VGG-2 Public Notebook link:
   https://www.kaggle.com/code/tenffe/vggface-cv-focal-loss
3. VGG-3 Public Notebook link:
   https://www.kaggle.com/code/mattemilio/smile-best-who-smile-last
4. VGG-4 Public Notebook link:
   https://www.kaggle.com/code/ranjitkumar1/kin-detection
5. VGG-5 Public Notebook link:
   https://www.kaggle.com/code/vaishvik25/blend-of-smiles/notebook